\pdfoutput=1

\documentclass[11pt]{article}

\usepackage[final]{acl}

\usepackage{times}
\usepackage{latexsym}
\usepackage{multirow}
\usepackage{booktabs}
\usepackage{amsmath,amsfonts}
\usepackage{algorithmic}
\usepackage{algorithm}
\usepackage{graphicx}
\usepackage{bm}
\usepackage{amsmath}

\usepackage[T1]{fontenc}

\usepackage[utf8]{inputenc}

\usepackage{microtype}

\usepackage{inconsolata}

\title{FSMR: A Feature Swapping Multi-modal Reasoning Approach with Joint Textual and Visual Clues}

\author{Shuang Li$^{*}$, Jiahua Wang$^{\dag}$, Lijie Wen$^{*}$ \\
        $^{*}$Tsinghua University \\ 
        $^{\dag}$Beijing Institute of Technology \\ 
        \texttt{lisa18@mails.tsinghua.edu.cn}}

\begin{document}
\maketitle
\begin{abstract}
Multi-modal reasoning plays a vital role in bridging the gap between textual and visual information, enabling a deeper understanding of the context. This paper presents the Feature Swapping Multi-modal Reasoning (FSMR) model, designed to enhance multi-modal reasoning through feature swapping. FSMR leverages a pre-trained visual-language model as an encoder, accommodating both text and image inputs for effective feature representation from both modalities. It introduces a unique feature swapping module, enabling the exchange of features between identified objects in images and corresponding vocabulary words in text, thereby enhancing the model's comprehension of the interplay between images and text. To further bolster its multi-modal alignment capabilities, FSMR incorporates a multi-modal cross-attention mechanism, facilitating the joint modeling of textual and visual information. During training, we employ image-text matching and cross-entropy losses to ensure semantic consistency between visual and language elements. Extensive experiments on the PMR dataset demonstrate FSMR's superiority over state-of-the-art baseline models across various performance metrics.The code is available at https://github.com/THU-BPM/FSMR.

\end{abstract}

\section{Introduction}
With the rise of social media, online news, and other multimedia platforms, textual information often coexists with information from other modalities like images. Multi-modal information processing has emerged as a crucial research direction in the field of natural language processing, regarded as a foundational and long-term task in both academia and industry \citep{yu2021ernie,chen2020uniter}.

\begin{figure*}[htbp]
    \centering
    \includegraphics[width=0.78\linewidth]{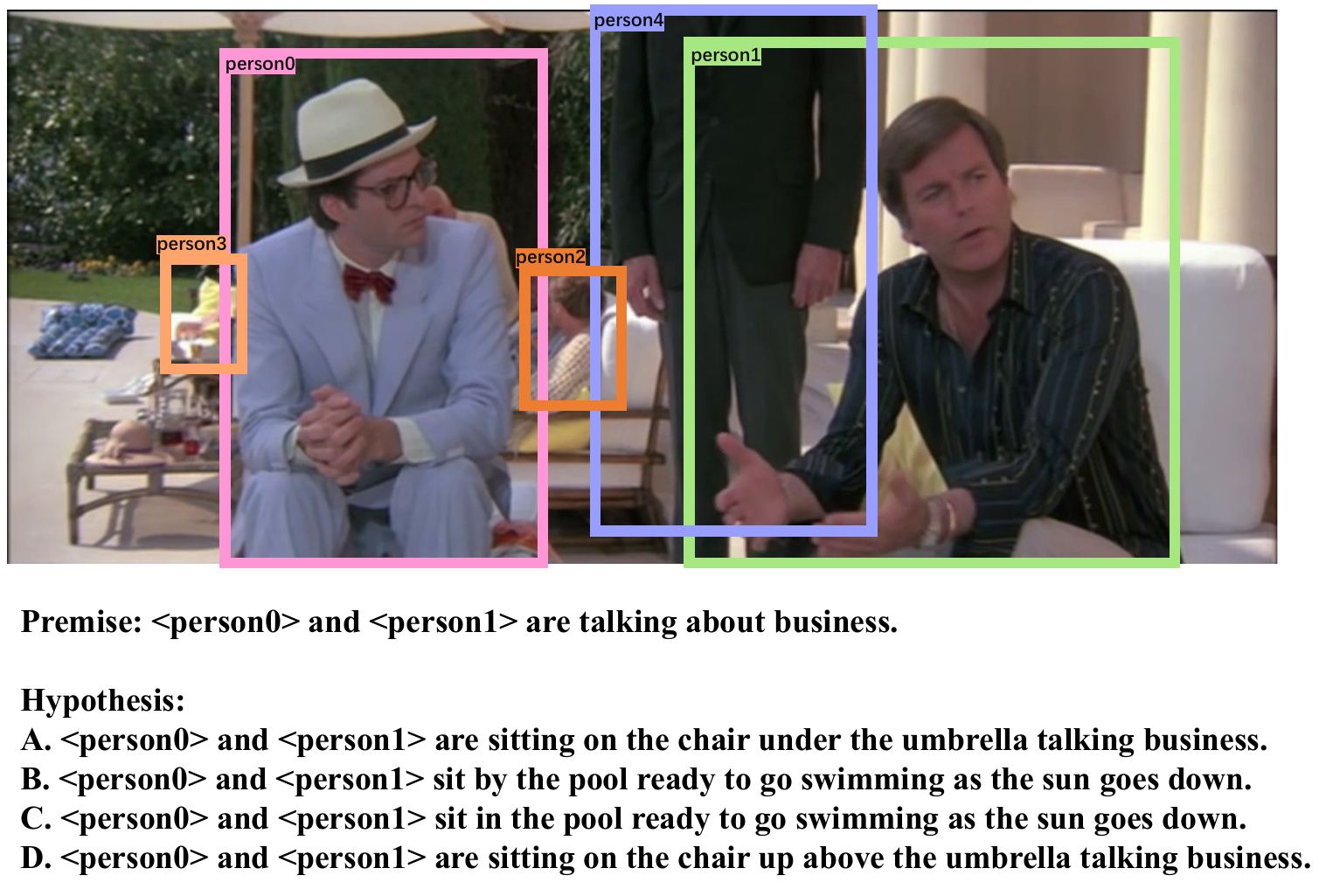}
    \caption{An example from PMR dataset}
    \label{fig:pmr-case}
\end{figure*}

Inspired by visual commonsense reasoning and textual inference, \citet{dong2022premise} introduced the PMR (Premise-based Multi-modal Reasoning) dataset. In this task, models are required to use textual information (from the premise) and visual cues (from the image) to infer whether the hypothesis is true or not. Figure \ref{fig:pmr-case} illustrates an example from the PMR dataset.
In this example, the model should recognize from the image that <person0> and <person1> are sitting together and conversing. Based on the textual premise, ``<person0> and <person1> are talking about business'', the model needs to determine whether the four hypotheses are true or not. 

Compared to pure textual reasoning, multi-modal reasoning is more complex because it requires models to establish deep semantic connections between various modalities.
To better fuse multi-modal inputs, researchers have designed self-supervised learning frameworks based on multi-modal encoders \cite{alberti2019fusion,li2019visualbert,tan2019lxmert}.
In recent years, given the outstanding performance of Pre-trained Language Models (PLM) in the field of natural language processing, many researchers have shown significant interest in Visual-Language Models (VLM) \cite{li2020oscar}. Although these methods have shown promise in reasoning tasks that heavily rely on visual cues, they still face challenges in aligning multi-modal data. For example, textual descriptions may zoom in on specific details of a scene, while the corresponding image may present an overall view of that scene.
These differences can make it difficult for models to effectively merge information from two modalities when performing reasoning tasks.
To address this problem, \citet{li-etal-2023-multi-modal} designed a multi-modal contextual reasoning framework. Unlike traditional models, this framework incorporates a prefix for aligning images with text in pre-trained language models, enabling context semantic learning for both language and vision.
However, the mentioned approaches do not delve into the fine-grained fusion of words in the premise and hypothesis with objects in the image, lacking granularity in multi-modal information integration.

Our paper introduces a Feature Swapping Multi-modal Reasoning (FSMR) model for multi-modal reasoning. The model utilizes a pre-trained visual-language model as an encoder, taking both text and image inputs to effectively represent features from both modalities. FSMR employs a unique feature swapping module that swaps the features of identified objects in the image with corresponding vocabulary words in the text, such as <person0> and <person1> in Figure \ref{fig:pmr-case}. The swapped features are then incorporated into a prompt template and input into the language model, allowing the model to understand the context information fused between images and text.
To further enhance the multi-modal alignment capability, FSMR introduces a multi-modal cross-attention mechanism, enabling joint modeling of textual and visual information. We adopt image-text matching loss and cross-entropy loss during training to ensure semantic consistency between vision and language. Extensive experiments on the standard PMR dataset for multi-modal reasoning validate the approach. Results demonstrate that FSMR outperforms state-of-the-art baseline models across various performance metrics.

The main contributions of this paper can be summarized as follows:
(1) We introduce the Feature Swapping Multi-modal Reasoning (FSMR) model to address multi-modal reasoning tasks; (2) We introduce a multi-modal cross-attention mechanism that allows for joint modeling of textual and visual information; (3) To substantiate our approach, we conducted extensive experiments on the PMR dataset. These experiments clearly demonstrate that FSMR surpasses state-of-the-art baseline models across various performance metrics.

\section{Related Work}
To better fuse multi-modal inputs, researchers have designed self-supervised learning frameworks based on multi-modal encoders. Specifically, depending on the construction of the encoder, these multi-modal learning frameworks can be categorized into two types. The first framework uses a unified encoder to directly process multi-modal inputs \cite{sun2019videobert,alberti2019fusion,li2019visualbert}. The second one initially employs two separate encoders to process textual and image data independently and then uses a joint encoder to integrate the representations obtained from both, achieving the goal of merging multi-modal information \cite{lu2019vilbert,tan2019lxmert}. Among these, \citet{cui-etal-2020-unsupervised} introduced a multi-modal alignment contrastive learning decoupled network. This approach introduces multi-modal contrastive losses between the text encoder and the image encoder, ensuring a high semantic match between the textual description and the corresponding image.

In recent years, given the outstanding performance of Pre-trained Language Models (PLMs) in the field of natural language processing, many researchers have shown significant interest in Visual-Language Models (VLMs) \cite{krojer2022image,li2020oscar,wang2022ofa}. \citet{lu2019vilbert} introduced a pre-trained model called VL-BERT for visual-language tasks. This model extends the Transformer encoder to accept visual and textual features as inputs. Still, in this process, context learning based on the multi-modal semantics of language and vision is often overlooked.
To address this problem, \citet{li-etal-2023-multi-modal} designed a multi-modal contextual reasoning framework. However, the mentioned approaches do not delve into the fine-grained fusion of words in the premise and hypothesis with objects in the image, lacking granularity in multi-modal information integration. In this paper, we proposes a Feature Swapping Multi-modal Reasoning (FSMR) model for multi-modal reasoning to tackle this problem.

\section{Architecture}
\begin{figure*}[htbp]
    \centering
    \includegraphics[width=0.8\linewidth]{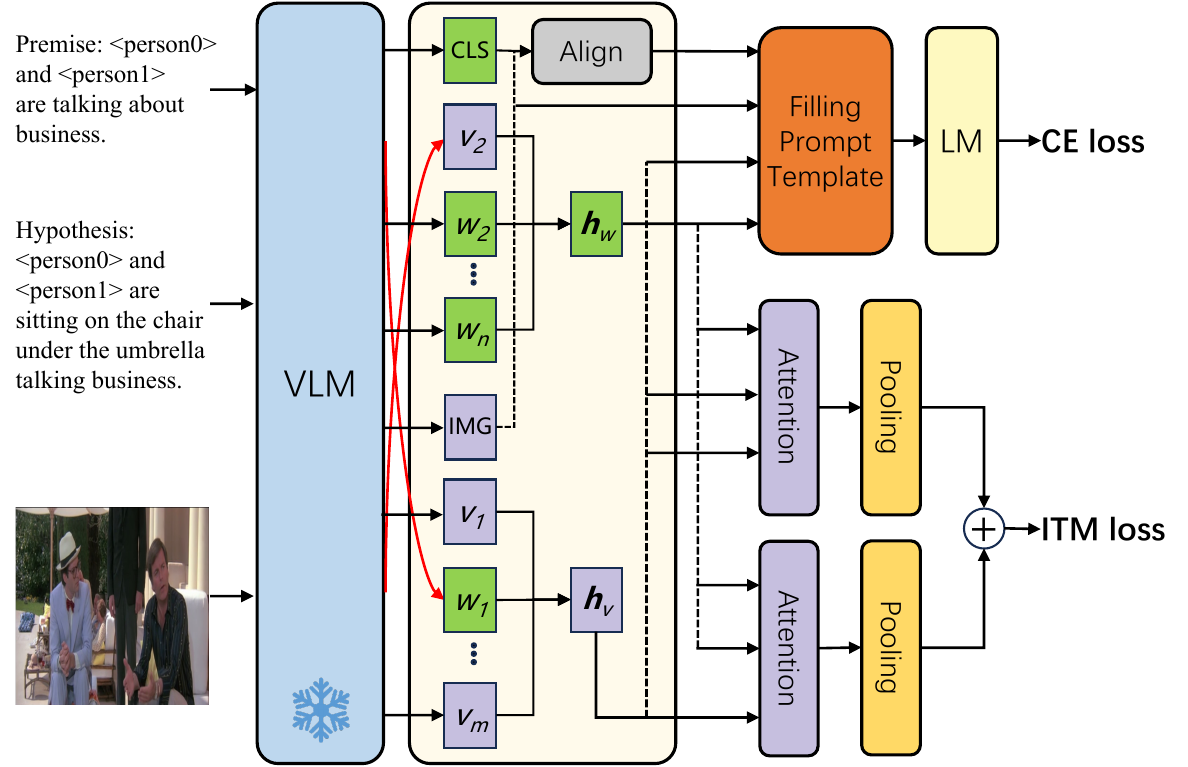}
    \caption{Overall Architecture of the FSMR Model}
    \label{fig:fsmr}
\end{figure*}
The overall structure of the FSMR model is depicted in Figure \ref{fig:fsmr}. The FSMR model utilizes a pre-trained Visual Language Model as an encoder to obtain representations of text and images. We introduce a feature swapping layer that swaps the features of objects in the image with corresponding word representations in the text. After obtaining these new representations, they are filled into a pre-designed prompt template and fed into a language model to compute cross-entropy loss. Additionally, FSMR incorporates a multi-modal multi-head attention module to integrate information from both text and images. The model employs an image-text matching loss to align text and image representations in the semantic space.
\subsection{Encoder}
In the PMR taspremise-based multi-modal reasoning task, each instance consists of two sentences (premise and hypothesis), an image denoted as $V$, and a label representing the relationship between the sentences (entailment or contradiction). The image input is denoted as $V$. An instance in a batch, denoted as $\mathcal{I}$, is represented as $(X^{(p)}, X^{(h)}, V, y)_i$, where $i = \{1, \dots, K\}$ is the sample index, and $K$ is the batch size. The goal is to learn a mapping function $f$ on the training data, which predicts the category $y$ based on the input.

The FSMR model utilizes the pre-trained ViLBERT as its encoder. To process complex inputs that combine text and images, FSMR employs a special concatenation method. The format is as follows: ``\texttt{[CLS]} $X^{(p)}$ \texttt{[SEP]} $X^{(h)}$ \texttt{[IMG]} $V$''.
The embeddings for text and images are obtained from the encoder:
\begin{equation}
\bm{h}_{\text{CLS}}, \bm{w}_i, \bm{v}_j, \bm{h}_{\text{IMG}} = \operatorname{ViLBERT}(X^{(p)}, X^{(h)},V)
\end{equation}
Where $\{\bm{w}_i | i=1,2,\cdots,n\}$ represents word representations in the premise and hypothesis. , with $n=l_1+l_2$. $\{\bm{v}_j | j=1,2,\cdots,m\}$ represents objects in the image, with $m$ denoting the number of objects. For \texttt{[CLS]} and \texttt{[IMG]}, the encoder outputs are denoted as $\bm{h}_{\text{CLS}}$ and $\bm{h}_{\text{IMG}}$, respectively, representing the overall representations for text and image. 

\subsection{Feature Swapping Layer}
\begin{figure}[htbp]
    \centering
    \includegraphics[width=0.9\linewidth]{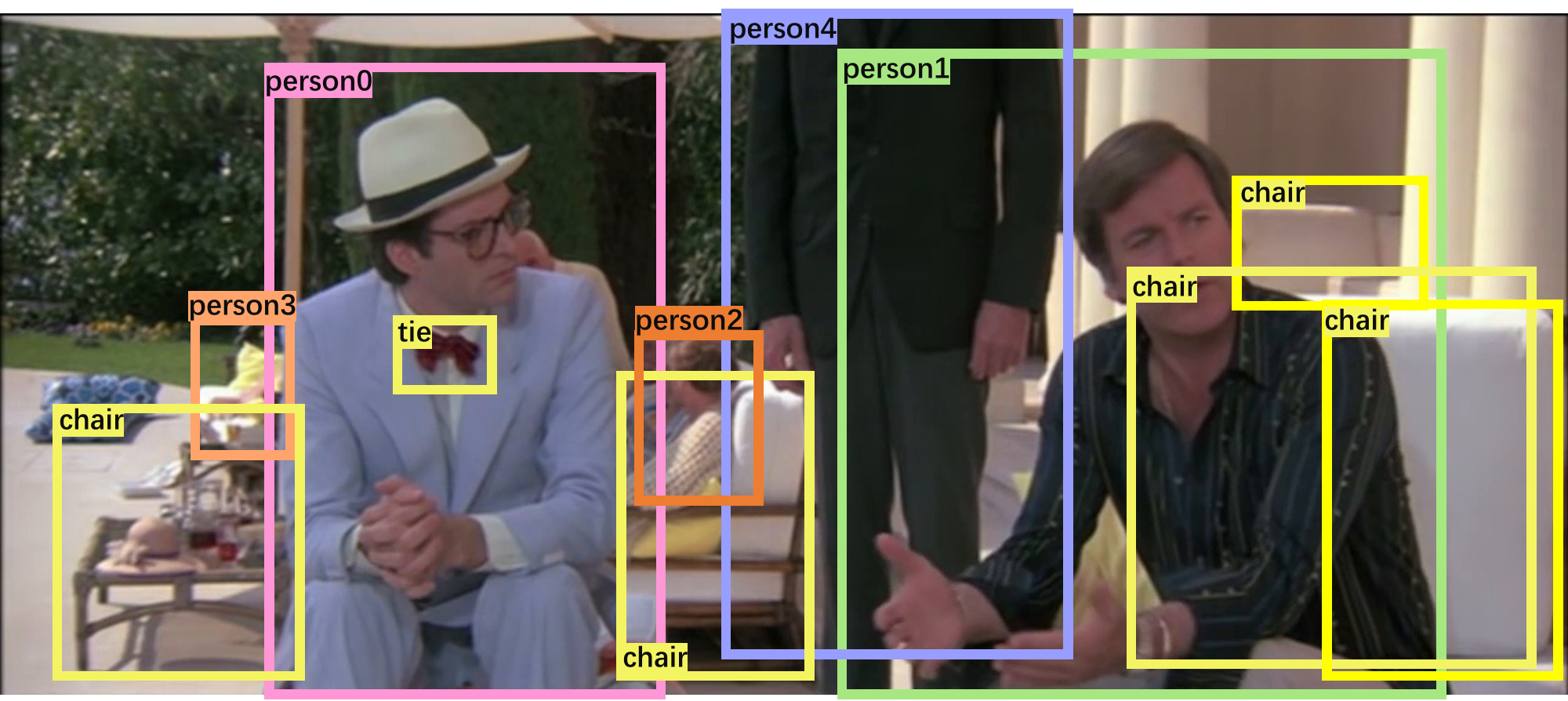}
    \caption{Example of objects in the image}
    \label{fig:obj}
\end{figure}

In order to enhance the model's understanding of multi-modal contexts, we introduce an innovative mechanism called Feature Swapping. For each object in the image, when a corresponding word is mentioned in the text, the embeddings representing that object in the image and the corresponding word in the text are swapped. Figure \ref{fig:obj} displays an image with objects outlined in boxes.
This image contains 11 objects (person0, person1, person2, person3, person4, tie, chair, chair, chair, chair, chair). Both ``person0'' and ``person1'' are mentioned in both the premise and hypothesis. 

For the example in Figure \ref{fig:fsmr}, the hypothesis describes that ``person0'' and ``person1'' are discussing business, which aligns with the meaning described in the premise. The goal of the Feature Swapping Layer is to ensure that the model correctly aligns words in the text with corresponding image objects in the semantic space. The exchanged embeddings for word representations $\{\bm{w}_i|i=1,2,\cdots,n\}$ and object representations $\{\bm{v}_j|j=1,2,\cdots, m\}$ are denoted as $\bm{h}_w$ and $\bm{h}_v$, respectively:
\begin{align}
    \bm{h}_w = (\bm{w}_1,\cdots,\bm{w}_{i-1},[\bm{v}]_j,\bm{w}_{i+1},\dots,\bm{w}_n)\\
\bm{h}_v = (\bm{v}_1,\cdots,\bm{v}_{j-1},[\bm{w}]_i,\bm{v}_{j+1},\cdots,\bm{v}_m)
\end{align}
In the above equations, $[\bm{v}]_j$ and $[\bm{w}]_i$ represent the swapped features $\bm{v}_j$ and $\bm{w}_i$, respectively, and their corresponding words and objects actually represent the same entity.

For the overall text representation $\bm{h}_{\text{CLS}}$ and the overall image representation $\bm{h}_{\text{IMG}}$, FSMR designs an aligner module to tightly integrate them, forming a fused representation of the image and text. This aligner module is not complex, consisting of linear layers and the Tanh activation function. The fused representation $\bm{A}$ is calculated by:
\begin{equation}
    \bm{A} = \operatorname{tanh}(\bm{W} * \operatorname{concat}(\bm{h}_{\text{CLS}}, \bm{h}_{\text{IMG}}) + \bm{b})
\end{equation}
$\bm{W}$ and $\bm{P}$ are trainable parameters used for linear transformation.

\subsection{Prompt Template}
After obtaining the aforementioned embedding representations, this section employs the widely adopted technique of prompt engineering to integrate the encoded information and fill it into a carefully designed prompt template. 
The predefined prompt template is as follows: ``\texttt{[CLS]} Given an image with feature <$\bm{h}_{\text{IMG}}$>, the alignment feature is <$\bm{A}$>, objects identified as <$\bm{h}_v$> \texttt{[SEP]} <$\bm{h}_w$>''. 

This template is then input into a pre-trained language model (RoBERTa \cite{liu2019roberta}). By constructing such prompt templates, existing image representations are embedded into the language model, transforming the multi-modal reasoning task paradigm into a purely language model reasoning paradigm. The output of RoBERTa's \texttt{[CLS]} representation, denoted as $\bm{S}_{\text{CLS}}$, is used for inference.

\subsection{Multi-Head Attention Module}
\label{sec:attn}

To effectively fuse language and visual information, this model introduces a multi-head attention module after the feature fusion layer.
Given intermediate representations for vision and language, denoted as $\bm{h}_v$ and $\bm{h}_w$ respectively, separate linear layers are used to compute the query, key, and value matrices. In the traditional way, the query, key, and value matrices all originate from the same input. However, we adopt a cross-modal multi-head attention mechanism in FSMR. Specifically, the key and value matrices for the language modality are provided to the multi-head attention component for the vision modality as input, and vice versa, the key and value matrices for the vision modality are provided to the multi-head attention component for the language modality. The representations output by the two multi-head attention components are denoted as $\bm{O}_w$ and $\bm{O}_v$, computed as follows:
\begin{align}
    \bm{O}_w = \text{Multi-Head}(Q_w, K_v, V_v)\\
    \bm{O}_v = \text{Multi-Head}(Q_v, K_w, V_w)
\end{align}

After obtaining the outputs of the two multi-head attentions, dimension reduction and capture of their main features are achieved first through a pooling layer (e.g., average pooling or max pooling). Let $\bm{P}_w$ and $\bm{P}_v$ be the representations of $\bm{O}_w$ and $\bm{O}_v$, respectively, after pooling processing:
\begin{align}
    \bm{P}_w = \text{Pooling}(\bm{O}_w)\\
    \bm{P}_v = \text{Pooling}(\bm{O}_v)
\end{align}

Next, the two pooled representations are concatenated to obtain the overall multi-head attention representation $\bm{S}_{\text{attn}}$:
\begin{equation}
\bm{S}_{\text{attn}} = \text{Concat}(\bm{P}_w, \bm{P}_v)
\end{equation}
$\text{Pooling}$ represents a pooling function (e.g., average pooling, max pooling, etc.), and $\text{Concat}$ denotes the vector concatenation operation.

\subsection{Objective Function}
\paragraph{Image-Text Matching Loss}
In this section, we introduce the image-text matching loss function specifically designed for FSMR, aiming to ensure the effective alignment of visual and textual information. This loss function is denoted as $\mathcal{L}_{\text{ITM}}$. After obtaining the overall representation $\mathbf{S}_{\text{attn}}$ from multi-head attention, it is first passed through a linear layer. Subsequently, it is transformed into a probability $p_{\text{ITM}}$ within the range of [0,1] using the sigmoid activation function:
\begin{equation}
   p_{\text{ITM}} = \text{sigmoid}(\mathbf{W}\cdot\mathbf{S}_{\text{attn}} + \mathbf{b})
\end{equation}
$\mathbf{W}$ and $\mathbf{b}$ are trainable parameters, and $y$ is the ground truth label for the example. Next, we calculate the loss function $\mathcal{L}_{\text{ITM}}$ as follows:
\begin{equation}
\mathcal{L}_{\text{ITM}} = -\left( y \log p_{\text{ITM}} + (1-y) \log(1 - p_{\text{ITM}}) \right)
\end{equation}

\paragraph{Cross-Entropy Loss}
In addition to the image-text matching loss, the \texttt{[CLS]} representation $\bm{S}_{\text{CLS}}$ generated by RoBERTa utilizes a softmax-based Cross-Entropy loss function for classification:

\begin{equation}
  \mathcal{L}_{\text{CE}} = CrossEntropy(\bm{W}\cdot\bm{S}_{\text{CLS}} + \bm{b}, y)
\end{equation}
$\bm{W}$ and $\bm{b}$ are trainable parameters, and $y$ represents the annotated label for this example.
\paragraph{Overall Loss Function}
The overall training objective of the FSMR model, denoted as $\mathcal{L}$, is the weighted average of the cross-entropy loss and the image-text matching loss, represented as:
\begin{equation}
  \mathcal{L} = \alpha\mathcal{L}_{\text{CE}} + \beta\mathcal{L}_{\text{ITM}}\label{fsmr-eq}
\end{equation}
$\alpha$ and $\beta$ are hyperparameters used to balance the loss functions.

\section{Experimental Setup}

\subsection{Benchmark Dataset}
To validate the effectiveness of the proposed model, experiments were conducted on the high-quality PMR dataset \cite{dong2022premise}. These samples were created through a multi-stage crowd-sourcing process. Crowd-workers, guided by predefined categories, selected high-quality movie screenshots and manually curated premise templates to write a genuine hypothesis along with three distractor options in a cross-checking procedure, based on the provided premise and the image. Classification accuracy is used as the evaluation metric in the experiments.

\subsection{Implementation Details}
The model is implemented using PyTorch. We utilize Faster R-CNN \cite{he2017mask} as the image feature encoder for extracting visual regions. For visual-linguistic alignment, we employ Oscar as the visual language aligner, and RoBERTa serves as the multi-modal context network. The training details can be found in Appendix \ref{sec:training details}.

\subsection{Baseline Models}
We compare FSMR with pre-trained language models and multi-modal models as follows:(1)BERT \cite{devlin2019bert} and RoBERTa \cite{liu2019roberta} are large-scale language models based on the Transformer architecture; (2)ViLBERT \cite{lu2019vilbert} is a cross-modal pre-trained model with dual data streams; (3)ERNIE-VL \cite{yu2021ernie} uses a single-stream fusion encoder and leverages structured knowledge obtained to learn joint representations; (4)UNITER \cite{chen2020uniter} integrates visual information and utilizes joint multi-modal embeddings to support heterogeneous downstream visual language tasks; (5) Oscar \cite{li2020oscar} is a single-stream fusion encoder model that simplifies alignment learning by using object labels detected in images as anchors; (6) OFA \cite{wang2022ofa} is a sequence-to-sequence cross-modal learning framework that unifies various cross-modal and uni-modal tasks; (7) MVPTR \cite{li2022mvptr} is a pre-trained cross-modal model that introduces multi-level semantic alignment between vision and language; (8) CALeC \cite{yang2022chunk} is a unified prediction and generation model for certain visual-language tasks; (9) PromptFuse \cite{liang2022modular} is a prompt-based learning approach to incorporate visual information into language models; (10) ModCR \cite{li-etal-2023-multi-modal} is a multi-modal contextual reasoning framework that incorporates a prefix capable of learning alignment between images and text into pre-trained language models.

\section{Experiment Results}
\subsection{Main Results}

\begin{table}[htbp]
  \centering
    \begin{tabular}{lcc}
    \toprule
    \textbf{Method} $\downarrow$ \textbf{Types} $\rightarrow$  & \textbf{Validation} & \textbf{Testing} \\
    \midrule
    BERT-B & - & 65.2 \\
    VL-BERT-B & - & 75.4 \\
    ERNIE-VL-B & - & 79.0 \\
    UNITER-B & - & 77.4 \\
    Oscar-B & 77.7 & 76.1 \\
    RoBERTa-L & 77.3 & 75.0 \\
    PromptFuse & 77.4 & 76.5 \\
    VL-BERT-L & - & 79.3 \\
    ERNIE-VL-L & - & 79.9 \\
    UNITER-L & - & 77.0 \\
    OFA-L & 79.9 & 79.1 \\
    MVPTR & 79.5 & 78.9 \\
    CALeC  & 80.1 & 78.7 \\
    ModCR & \underline{85.0} & \underline{83.6} \\
    FSMR & \textbf{86.4} & \textbf{84.8} \\
    \bottomrule
    \end{tabular}%
    \caption{Model Performance (accuracy) on the PMR dataset. The results of BERT, VL-BERT, ERNIE-VL and UNITER are reported by \citet{dong2022premise}. For baselines, ``B'' and ``-L'' indicate the base and large version, respectively. The underscore and bold indicate the second highest value and best performance(same as following tables).}
  \label{tab:result-pmr}%
\end{table}%

We conducted experiments on the PMR dataset to evaluate the model's performance. Table \ref{tab:result-pmr} displays the results of FSMR and other baseline models on both the validation and test sets. All results are the averages of five runs with different random seeds, and the best results are highlighted in bold. Some models, such as BERT-B, VL-BERT-B, ERNIE-VL-B, VL-BERT-L, and UNITER-L, were evaluated only on the test set, and validation set data were not provided in the original work.

From the test set data, it is evident that most models perform in the range of 75\% to 80\%. This demonstrates that the multi-modal natural language reasoning task on the PMR dataset is indeed challenging. FSMR excels, achieving the best performance on both the validation and test sets, with accuracy rates of 86.4\% and 84.8\%, respectively, significantly outperforming other baseline models. Compared to the state-of-the-art baseline model ModCR, FSMR exhibits a substantial improvement, increasing accuracy by 1.4\% on the PMR validation set and 1.2\% on the test set. This improvement is relatively significant in natural language processing tasks.
The performance of BERT-B and RoBERTa (text input only) suggests that reasoning based solely on the premise text can lead to correct choices, but with lower accuracy. 
FSMR, using RoBERTa-L as its primary backbone, outperforms pre-trained VLM and LM models on both datasets. This indicates that the FSMR approach effectively integrates semantic information from different modalities when performing inference.

\begin{table}[ht]
\centering

\begin{tabular}{lcccc}
\toprule
\textbf{Method} $\downarrow$ \textbf{Types} $\rightarrow$ & \textbf{AT}$\uparrow$ & \textbf{D1}$\downarrow$ & \textbf{AF}$\downarrow$ & \textbf{D2}$\downarrow$ \\
\midrule
BERT-B & 65.2 & 19.8 & 19.6 & 4.5 \\
Oscar-B & 76.1 & 10.2 & 12.1 & 1.7 \\
RoBERTa-L & 75.0 & 17.7 & 6.1 & \underline{1.2} \\
PromptFuse & 76.5 & 16.5 & \underline{5.9} & \underline{1.2} \\
ERNIE-VL-L & 79.9 & 10.7 & 8.2 & \underline{1.2} \\
OFA-L & 79.1 & 9.7 & 9.9 & 1.3 \\
MVPTR & 78.9 & \textbf{7.5} & 11.8 & 1.8 \\
CALeC & 78.7 & 8.6 & 10.9 & 1.8 \\
ModCR & \underline{83.6} & 9.2 & \textbf{5.6} & 1.6 \\
FSMR & \textbf{84.8} & \underline{8.4} & \underline{5.9} & \textbf{0.9} \\
\bottomrule
\end{tabular}
\caption{Detailed performance on the test set of PMR. The results of BERT and ERNIE-VL are reported by \cite{dong2022premise}. AT, D1, AF, D2 represent the Action True and Image True, Action True yet Image False, Action False yet Image True, Action False and Image False, respectively. ``Action True or False'' indicate the answer whether meets the premise. Similarly, ``Image True or False'' show the answer whether meets the image information.
}
\label{tab:test-pmr}%
\end{table}
Table \ref{tab:test-pmr} provides a comprehensive overview of the model's performance on the PMR test set, aiming to evaluate the model's accuracy in reasoning across different types of answer candidates. The table presents the model's reasoning distribution across these categories, allowing for an in-depth analysis of potential factors contributing to classification errors—whether they are due to semantic disparities or deviations in image information.
Observing the table, FSMR exhibits superior overall performance compared to other baseline models, with error rates of 8.4\%, 5.9\%, and 0.9\% in D1, AF, and D2, respectively. Particularly in the D2 category, FSMR outperforms all other models. By combining RoBERTa as the context encoder for prompts, FSMR successfully achieves precise alignment of textual and image semantics through feature swapping and cross-modal multi-head attention mechanisms. This not only retains robust text reasoning abilities, as indicated by the AF results, but also significantly enhances the utilization of image information, as shown in the D1 results. In summary, adding a visual-language semantic alignment mechanism to vision-augmented language models is crucial. Moreover, there is still room for optimization in the area of contextual reasoning in current Vision-Language Models.

\subsection{Ablation Study}

\begin{table}[htbp]
  \centering
    \begin{tabular}{lcc}
    \toprule
    \textbf{Model}   & \textbf{Validation} & \textbf{Testing} \\
    \midrule
    FSMR & \textbf{86.3} & \textbf{84.8} \\
    -Feature Swapping & 85.8 & 84.4 \\
    -Prompt Template & 84.5 & 83.4 \\
    -Multi-head Attention & 85.3 & 83.9 \\
    -ITM loss & 85.5 & 84.1 \\
    -CE loss & 85.4   & 84.2 \\
    \bottomrule
    \end{tabular}%
    \caption{The ablation results of FSMR on the test set of PMR.}
  \label{tab:fsmr-ablation}%
\end{table}%

To gain a better understanding of the contributions of each key component within FSMR, we conducts ablation studies on the PMR dataset.
The results are presented in Table \ref{tab:fsmr-ablation}. Notably, removing the feature swapping layer results in a performance drop of 0.5\% on the validation set and 0.4\% on the test set, emphasizing its importance in enhancing alignment between image objects and textual words. Removing the prompt template has a more significant impact, causing accuracy to decrease by 1.8\% on the validation set and 1.4\% on the test set. This demonstrates that the prompt template plays a crucial role in incorporating image information into the language context, which is vital for reasoning accuracy.
The removal of the multi-head attention module leads to a substantial performance drop, with accuracy decreasing by 1.0\% on the validation set and 0.9\% on the test set. This highlights the critical role of the multi-head attention module in aligning and fusing textual and visual information effectively.
Removing the image-text matching loss alone results in a decrease of 0.8\% on the validation set and 0.7\% on the test set, underscoring its positive impact on training the model to align image and text information.
Finally, the removal of the cross-entropy loss has a relatively smaller impact, causing a decrease of 0.9\% on the validation set and 0.6\% on the test set.

\subsection{Analysis of Feature Swapping}

\begin{table}[htbp]
  \centering

    \begin{tabular}{lcc}
    \toprule
    \textbf{Method} & \textbf{Val} & \textbf{Test} \\
    \midrule
    Unidirectional (Image to Text) & 85.9 & 84.4 \\
    Unidirectional (Text to Image) & 84.7 & 83.8 \\
    Bidirectional & \textbf{86.3} & \textbf{84.8} \\
    Hybrid & 85.4 & 83.7 \\
    \bottomrule
    \end{tabular}%
    \caption{Experimental Results with Different Feature Swapping Strategies}
  \label{tab:fsmr-swap}%
\end{table}%

Table \ref{tab:fsmr-swap} illustrates the experimental results of different feature swapping methods within the Feature Swapping Layer. When replacing text features with image features, the model achieves validation and test set accuracy of 85.9\% and 84.4\%, respectively, which exhibit a relatively modest decrease compared to bidirectional swapping. However, when replacing image features with text features, the model's performance is notably lower, with validation and test set accuracy of 84.7\% and 83.8\%, respectively.

The model performs exceptionally well with bidirectional feature swapping, achieving validation and test set accuracy of 86.3\% and 84.8\%, respectively. 
In the case of hybrid swapping, which involves randomly choosing one of the four methods (unidirectional image, unidirectional text, bidirectional, or no swapping), the model's performance is slightly lower than bidirectional swapping but falls between the two unidirectional methods. The accuracy on the validation and test sets for hybrid swapping is 85.4\% and 83.7\%, respectively, indicating that the hybrid swapping strategy indeed leverages some of the advantages of bidirectional swapping but may not consistently achieve optimal performance under all conditions.

\subsection{Analysis of Multi-Head Attention}

\begin{figure*}
    \centering
    \includegraphics[width=0.95\linewidth]{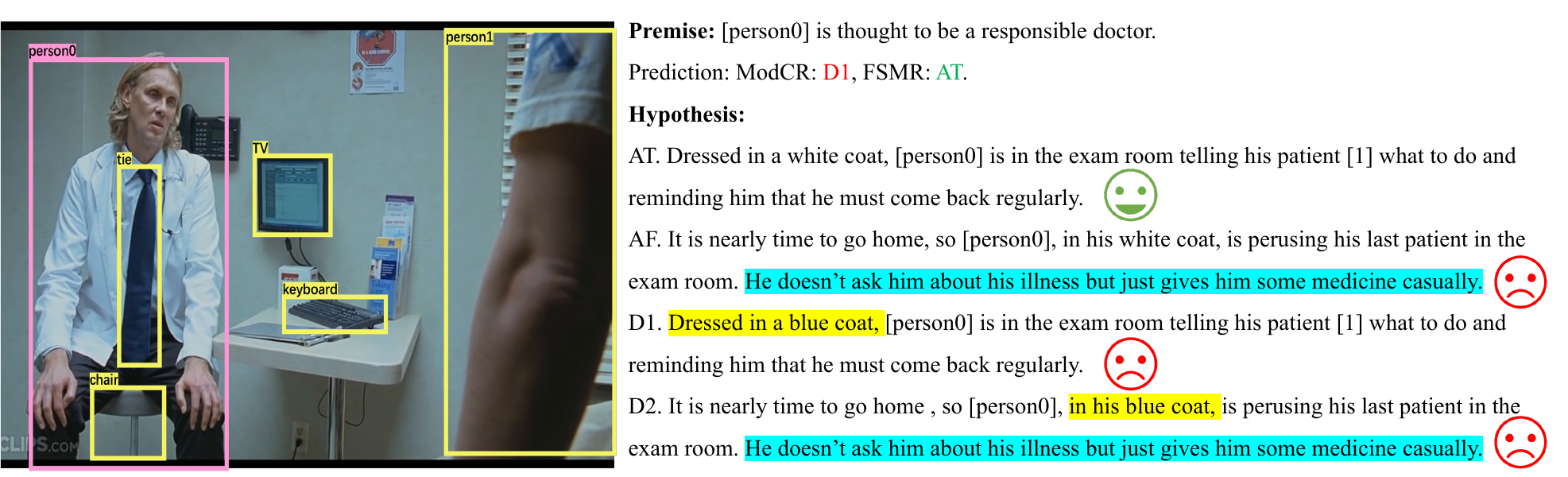}
    \caption{A case from the PMR Test Set. Blue indicates content that contradicts the textual premise, while yellow marks content inconsistent with the image. Green and red emoticons signify correct and incorrect options, respectively.}
    \label{fig:case}
\end{figure*}
\begin{table}[htbp]
  \centering
  
    \begin{tabular}{lcc}
    \toprule
    \textbf{Strategy} & \textbf{Validation} & \textbf{Testing} \\
    \midrule
    Visual Attention & 86.1 & 84.1 \\
    Language Attention & 84.5 & 82.8 \\
    Mixed Attention & \textbf{86.3} & \textbf{84.8} \\
    \bottomrule
    \end{tabular}%
    \caption{Experimental Results with Different Multi-Head Attention Strategies}
  \label{tab:fsmr-attn}%
\end{table}

As shown in Table \ref{tab:fsmr-attn}, this section analyzes the impact of different multi-head attention strategies on model performance.

From the table, it can be observed that when the model uses visual modality attention only, it achieves an accuracy of 86.1\% on the validation set and 84.1\% on the test set. In contrast, when using language modality attention only, the model's accuracy on the validation and test sets is 84.5\% and 82.8\%, significantly lower than the pure visual modality strategy. 
When both visual and language modality attention mechanisms are used simultaneously, the model's accuracy on the validation and test sets surpasses that of single modality strategies, reaching 86.3\% and 84.8\%, respectively. This demonstrates that combining visual and language information leads to better performance and underscores the importance of multi-modal attention in understanding and integrating modality information.

\subsection{Case Analysis}

A case from the PMR test set is illustrated in Figure \ref{fig:case}.
In this case, the textual premise states that "[person0] is considered a responsible doctor," and in the image, [person0] is seen wearing a white coat while sitting in a chair. Among the four options, the 'AT' option conveys that "[person0] is wearing a white coat, providing guidance in the examination room and reminding the patient to return for a follow-up," which aligns with both the image and the textual premise. The 'AF' option suggests that "[person0] is about to finish work, did not inquire about the patient's condition, and casually provided some medication," which contradicts the responsible doctor mentioned in the premise. However, in the 'D1' option, the blue coat contradicts the image information. The 'D2' option combines elements from both 'AF' and 'D1' and is inconsistent with both the textual and image information.

For this example, the baseline ModCR model's inference results in 'D1,' indicating that this model failed to effectively integrate image information for reasoning and did not recognize the contradiction between the answer and the image content. In contrast, FSMR can jointly model multi-modal information to infer the correct answer, identifying inconsistencies with both the image and textual premise. This demonstrates that FSMR, through multi-modal attention mechanisms and alignment loss, can fuse and comprehend textual and image data, enabling cross-modal contextual semantic reasoning.

\section{Conclusion}
We propose a Feature Swapping Multi-modal Reasoning model named FSMR. The features that are swapped are subsequently integrated into a prompt template and fed into a language model.

To further enhance the alignment and complementarity between text and images, FSMR introduces a multi-modal cross-attention mechanism, which plays a pivotal role in deepening the integration of visual and language information. Additionally, the model's training strategy is meticulously designed, ensuring that FSMR effectively aligns and integrates visual and textual information in the context of multi-modal reasoning tasks.
Experimental evaluations demonstrate FSMR's superior performance on the standard PMR dataset. Furthermore, we delves into a comprehensive exploration and analysis of the components of the FSMR model.

\section{Limitations}
The FSMR model exhibits promising advancements in multi-modal reasoning, but certain limitations should be considered. Its performance heavily relies on diverse and high-quality training data, and generalization to different domains beyond the PMR dataset may be a challenge. Additionally, while superior on the PMR dataset, FSMR's performance on other multi-modal datasets remains unexplored. Addressing these issues is crucial for enhancing the model's practical applicability across various multi-modal reasoning tasks.

\bibliography{anthology,custom}
\appendix

\section{Training Details}
\label{sec:training details}
During the model training process, we employ the RMSprop optimizer \cite{tieleman2012rmsprop}. We train the model for 30 epochs with a batch size of 8. The base learning rate of the model is set to 4e-06, with a weight decay of 8e-05, $\epsilon$ set to 5e-05, and it is adjusted using a linear scheduler. To ensure that the processed sequence information does not exceed the model's capacity, we set the maximum sequence length to 150. The length of the visual prefix is set to 3, while the cross-modal alignment prefix is set to 5. The number of heads in the multi-modal multi-head attention module in the model is set to 16, with a dropout rate of 0.2. All experiments are conducted 5 times using different random seeds, and the average results are reported. All methods select the best-performing model using the validation set. To ensure efficient computation, all experiments are carried out on GeForce GTX 3090Ti.

\end{document}